%% file: main.tex
\documentclass[runningheads]{llncs}

\usepackage{accv}

\usepackage{accvabbrv}

\usepackage{graphicx}
\usepackage{booktabs}
\usepackage{bm}
\usepackage{multirow}
\usepackage{makecell}

\usepackage[accsupp]{axessibility}  %

\usepackage[hidelinks]{hyperref}

\usepackage{orcidlink}

\newcommand{\YYY}{NarrativeFlow\xspace}
\newcommand{\DifferenceDescription}{\DD}
\newcommand{\DD}{\textcolor{black}{Narrative Delta}\xspace}
\newcommand{\ND}{\textcolor{black}{ND}}
\newcommand{\MultiXXX}{\textcolor{black}{subtask}\xspace}
\newcommand{\XXXFusion}{Vision-Language Fusion Encoder\xspace}
\newcommand{\xxxfusion}{vision-language fusion encoder\xspace}
\newcommand{\XXX}{\textcolor{black}{subtask}\xspace}
\newcommand{\PS}{\textcolor{black}{st}}

\newcommand{\figref}{Fig.~\ref}
\newcommand{\tabref}{Table~\ref}

\usepackage[most]{tcolorbox}
\newcommand{\TColorBox}[2]{%
\begin{tcolorbox}[
  colback=blue!5!white,
  colframe=blue!75!black,
  fonttitle=\bfseries,
  title=#1,
  boxrule=0.5pt,
  sharp corners,
  left=2mm, right=2mm, top=1mm, bottom=1mm,
  before upper={\let\\\par},
]
#2
\end{tcolorbox}
}

\begin{document}

\title{NarrativeFlow: \\ Flow-Based Vision-Language-Action Model \\ Using Robot Velocity Fields
} 

\titlerunning{NarrativeFlow: VLA Model using Robot Velocity Fields}

\author{Shota Kobayashi\and
Koki Seno\and
Daichi Yashima\and
Komei Sugiura}
\authorrunning{S.~Kobayashi et al.}

\institute{
Keio University, Japan
\email{\{shotakoba10267,koki.seno,ydaichi1207,komei.sugiura\}@keio.jp}
}

\maketitle

\input{section0}

\input{tab/1-10-eyecatch}

\input{section1}

\input{section2-1}

\input{section4-1}
\input{section5-2}

\input{section8}
\input{section7}

\input{section9}

\bibliography{reference}  %
\input{supp}

\bibliographystyle{splncs04}
\end{document}

%% file: section0.tex
\begin{abstract}
We focus on language-conditioned flow-based manipulation, where robot flows (robot velocity fields) serve as embodiment-agnostic, motion-centric representations for leveraging data collected from multiple robot platforms.
This task is crucial because language-conditioned manipulation is essential for practical robotic systems, yet scaling robot foundation models remains limited by the labor-intensive collection of embodiment-specific data.	
Existing methods either coarsely approximate robot flows with sparse keypoint displacements, or cannot handle language-conditioned manipulation.
To address this limitation, we propose \YYY, which models robot flows as continuous velocity fields using a flow-matching formulation conditioned on language.
Accordingly, \YYY generates robot flows that are physically consistent with real-world manipulation.
To validate \YYY, we have conducted experiments on standard datasets for language-conditioned manipulation.
The experimental results show that \YYY outperforms representative baseline methods on standard evaluation metrics.
Furthermore, through real-world experiments, we show that \YYY achieves higher success rates than baseline methods across multiple manipulation tasks.
The project page is available at \url{https://shota0520.github.io/NarrativeFlow-project-page/}

  \keywords{Learning from cross-embodiment data \and Flow matching \and Language-conditioned manipulation}
\end{abstract}

%% file: tab/1-10-eyecatch.tex
\begin{figure}[tb]
  \centering
  \includegraphics[width=\textwidth,height=6.5cm,keepaspectratio]{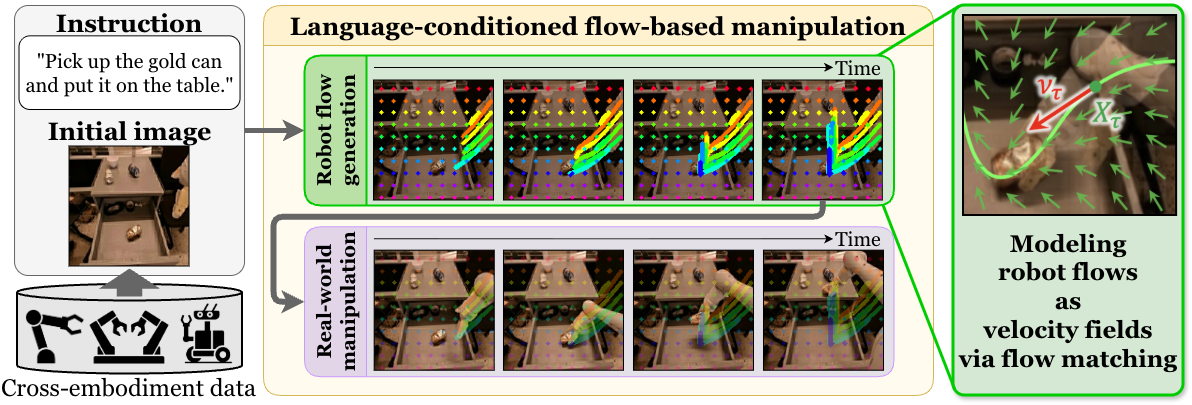}
  \caption{Overview of our approach to language-conditioned flow-based manipulation. We formulate robot flows as continuous velocity fields via flow matching. For training, we use data collected from multiple robot platforms. At test time, given an instruction and an initial image, our model generates a robot flow. The robot then executes the object manipulation task based on the robot flow.}
  \label{fig:eye-catch}
\end{figure}

%% file: section1.tex
\section{Introduction}
\label{sec:intro}
Labor shortages have increased demand for robotic systems that support workers and reduce physical burden in manufacturing, logistics, and healthcare.
To meet this demand, such systems must perform diverse tasks across a wide range of real-world environments, while following natural language instructions.
In this direction, recent work has explored robot foundation models, which are often pretrained on large-scale data collected from multiple robot platforms~\cite{pi0-black25rss, pi05-black25corl, oxe-neill24icra, openvla-kim24corl}.
This line of work indicates that the scale and diversity of training data are critical to model performance, yet training these models still requires labor-intensive collection of embodiment-specific data~\cite{pi0-black25rss, pi05-black25corl, oxe-neill24icra, openvla-kim24corl}.

In this study, we focus on language-conditioned flow-based manipulation.
In this task, a model generates a robot flow (robot velocity field) conditioned on an instruction, and then a robot performs object manipulation based on the robot flow. 
Robot flows serve as embodiment-agnostic intermediate representations that bridge visual observations and robot actions. 
Because such flows can be extracted from arbitrary robot videos, flow-based approaches can leverage large-scale data collected from multiple robot platforms rather than relying on costly embodiment-specific data collection.
However, this task remains challenging: indeed, a representative method, Im2Flow2Act~\cite{im2flow2act-xu24corl}, still underperforms a method using oracle robot flows by 21 percentage points in average success rate (see \tabref{tab:real-quantitative}).

Many existing flow-based manipulation methods~\cite{flip-gao25iclr, im2flow2act-xu24corl, track2act-bharadhwaj24eccv} do not formulate robot flows as dense velocity fields, but as displacements of sparse keypoints, even though robot manipulation is inherently a time-evolving physical process.
Specifically, these methods predict sparse keypoint coordinates and then derive robot flows from their displacements, yielding only a coarse approximation of continuous robot motion.
These formulations can be problematic because they assume that the displacements are deterministic, although such an assumption is rarely satisfied in practice.
By contrast, Flow as Flow~\cite{faf-seno26} can model robot flows as dense velocity fields; however, it cannot handle language-conditioned settings.
Moreover, existing methods~\cite{flip-gao25iclr, im2flow2act-xu24corl, track2act-bharadhwaj24eccv, faf-seno26} typically rely only on primary flow-generation objectives during training, which provide no explicit supervision for capturing task-relevant semantics.
As a result, such models can be misled by task-irrelevant visual details, such as background appearance and lighting.

To address these limitations, we propose \YYY, a language-conditioned robot flow generation method that formulates robot flows as probability velocity fields within a flow-matching framework.
By modeling velocity fields, \YYY generates robot flows that are more physically consistent with real-world manipulation.
Moreover, we introduce an auxiliary loss that aligns the model’s intermediate representations with task-relevant scene changes, enabling the model to capture task-relevant semantics.

\figref{fig:eye-catch} shows an overview of our approach.
We first train our model on cross-embodiment datasets. 
At deployment, given an instruction and an initial image, the model generates a robot flow that represents the robot motion required for the specified task (e.g., picking up the can and placing it on the table).
Subsequently, the robot executes the manipulation based on the generated flow to complete the task.

The contributions of this study are as follows:
\begin{enumerate}
    \item We propose \YYY, a language-conditioned robot flow generation method that formulates robot flows as probability velocity fields within a flow-matching framework, enabling high-quality robot flow generation.
    \item We introduce the \DifferenceDescription loss as an auxiliary objective that aligns the model’s intermediate representations with task-relevant scene changes. We further introduce two \MultiXXX tokens to mitigate interference between this auxiliary objective and the primary flow-generation objective.
    \item We show that \YYY outperforms baseline methods across standard datasets.
    \item Through real-world experiments, we show that \YYY achieves higher success rates than baseline methods across multiple manipulation tasks.
\end{enumerate}

%% file: section2-1.tex
\section{Related Work}

\noindent\textbf{\emph{Language-Conditioned Manipulation.}}
Robotic foundation models aim to address diverse tasks across varied environments and have been actively investigated~\cite{surveyA-kawaharazuka2025vla-survey, surveyB-zhong2025survey}.
Recent language-conditioned manipulation policies build on pretrained foundation models such as multimodal large language models (MLLMs), adopting them as semantic backbones coupled with action heads that map language and observations to control~\cite{pi0-black25rss,pi05-black25corl,evo1-lin2026cvpr}.
This design inherits semantic and perceptual priors from large-scale pretraining, which transfer to unseen tasks and objects~\cite{pi05-black25corl,instructvla-yang25iclr}.

A separate line of work introduces an explicit intermediate representation between language and action~\cite{im2flow2act-xu24corl,vidbot-chen25cvpr,ghost-krishna26rss}.
Such methods predict task- or scene-level targets, including 3D affordance maps (e.g.,~\cite{robobrain-ji25cvpr,ghost-krishna26rss}) and visual subgoals (e.g.,~\cite{susie-black24iclr,cotvla-zhao25cvpr}), rather than conditioning the policies directly on latent features from the backbones.
Among such targets, robot flows represent task progress as point or end-effector motion, bridging language specifications and the low-level trajectories that execute them~\cite{tramoe-yang25cvpr,im2flow2act-xu24corl,flip-gao25iclr,lilac-kambara26ral}.

\noindent\textbf{\emph{Flow-Based Manipulation.}}
Flow-based manipulation methods use robot flows as embodiment-agnostic motion representations that bridge vision and action~\cite{track2act-bharadhwaj24eccv,ec-flow-chen25iccv,g3flow-chen25cvpr,generalflow-yuan25corl,im2flow2act-xu24corl}. 
These representations can be extracted from heterogeneous robot videos (e.g.,~\cite{robotap-vecerik24icra,atm-wen23rss}) and from web-scale human videos (e.g.,~\cite{track2act-bharadhwaj24eccv,faf-seno26,egoscaler-yoshida25cvpr}).
Moreover, a large body of work has demonstrated the effectiveness of robot flows across diverse real-world tasks including rigid, articulated, and deformable object manipulation (e.g.,~\cite{flow-bot-pp-zhang23corl,im2flow2act-xu24corl}).

Although earlier methods generate robot flows over the entire frame (e.g.,~\cite{track2act-bharadhwaj24eccv,atm-wen23rss}), some studies use object-centric or hand-centric flows to further reduce the influence of embodiment-specific factors (e.g.,~\cite{im2flow2act-xu24corl,lilac-kambara26ral,ec-flow-chen25iccv,egoscaler-yoshida25cvpr}).
Other approaches adopt flow-conditioned video generation to predict the future scene (e.g.,~\cite{flip-gao25iclr}), or extend 2D flows into 3D using depth estimation (e.g.,~\cite{g3flow-chen25cvpr,ppi-yang25rss}).

These methods derive robot flows from the displacements of sparse keypoints. As an alternative, Flow as Flow~\cite{faf-seno26} directly predicts robot velocity fields via flow matching, a framework that predicts probability velocity fields in a generation space.
However, this method is limited to goal-conditioned manipulation and hence it is unable to address language-conditioned settings.
In contrast, \YYY generates robot flows using the flow-matching formulation conditioned on language.

\noindent\textbf{\emph{Learning Scene-Change Representations.}}
To model temporal dynamics and task-relevant transformations, prior work has leveraged scene-change information in visual representation learning.
For example, procedure-planning methods predict sequences of high-level action labels from a current visual observation toward a visual goal in instructional video understanding~\cite{pdpp-wang23cvpr,skipplan-li23iccv,kepp-nagasinghe24cvpr}.
Other work has focused on scene changes for object-centric localization (e.g.,~\cite{oscs-xue24cvpr}), procedural modeling (e.g.,~\cite{niu2024schema}), and fine-grained temporal modeling (e.g.,~\cite{hifvla-lin26cvpr,progress-xue25cvpr}).
In parallel, recent work has explored efficient temporal representations using video codecs, further leveraging temporal and scene-change cues (e.g.,~\cite{remora-yashima26cvpr}).

In manipulation tasks, goal images are often used to specify desired final scene configurations~\cite{rtsketch-sundaresan24corl, susie-black24iclr}.
These images provide fine-grained details of the scene change to be achieved, but are typically unavailable at inference time in language-conditioned settings.
To leverage this information for language-conditioned manipulation, GRIF~\cite{grif-myers23corl} uses initial--goal image pairs during training to learn scene-change representations aligned with instructions.

In addition, world models and dynamics models predict how objects and scenes change conditioned on actions.
They model such changes through future observation prediction, learned physical dynamics, or dense geometric motion~\cite{susie-black24iclr, graphdynamics-deng24icra, pointworld-huang26cvpr}.

%% file: section4-1.tex
\section{Method}
\input{section3}

\subsection{Overview}
\label{sec:overview}
We propose \YYY, a method for language-conditioned robot flow generation, inspired by flow-based manipulation policies~\cite{faf-seno26, flip-gao25iclr, im2flow2act-xu24corl}.
\input{tab/4-4-model}
\figref{fig:model} shows the architecture of \YYY.
Our model consists of the \xxxfusion and the Flow-as-Flow module.
The model inputs are an instruction $\bm{\ell}$ and an initial image $\bm{I}_0$, and they are encoded using Qwen3-Embedding-0.6B~\cite{qwen3embedding-zhang25} and DINOv3-ViT-B/16~\cite{dinov3-simeoni25} to obtain language and visual embeddings $\bm{z}_\text{lang}$ and $\bm{z}_\text{img}$, respectively.

\subsection{\XXXFusion}
\label{4-8a}
The \xxxfusion integrates $\bm{z}_\text{img}$ and $\bm{z}_\text{lang}$ into two purpose-specific joint representations, which are used in two downstream paths: one for conditioning the Flow-as-Flow module (Sec.~\ref{4-8c}) and the other for computing the \DD loss (Sec.~\ref{4-8b}).
A standard approach to aggregating multimodal inputs into a single representation is to add a learnable special token (e.g., a [CLS] token) to the inputs and use its output as a global readout~\cite{uniter-chen20eccv, lxmert-tan19}.
However, this single-token design can be insufficient for our model, in which the same token is used both to condition the Flow-as-Flow module and to compute the \DD loss.
This can entangle the learning signals from both downstream paths, hindering the token from capturing purpose-specific information.

To address this issue, we introduce two \MultiXXX tokens, each dedicated to a distinct downstream path.
This separation enables each token to act as a purpose-specific readout, helping to mitigate interference between the learning signals from the two downstream paths.
Specifically, the \XXX tokens are concatenated with $\bm{z}_\text{img}$ and $\bm{z}_\text{lang}$, and the resulting token sequence is processed by a Transformer encoder consisting of Transformer layers.
The output tokens corresponding to the \XXX tokens are projected into $\bm{z}_\text{\PS}^{(1)}$ and $\bm{z}_\text{\PS}^{(2)}$, which are used to condition the Flow-as-Flow module and to compute the \DD loss, respectively.

\subsection{\DifferenceDescription Loss}
\label{4-8b}
We introduce the \DD loss, which is an auxiliary loss, to guide $\bm{z}^{(2)}_\text{\PS}$ toward narrative representations~\cite{goko24corl} of task-relevant scene changes.
Existing robot flow generation methods~\cite{flip-gao25iclr,im2flow2act-xu24corl,faf-seno26} are typically trained using only a loss for flow generation (e.g., Eq.~\eqref{eq:primary_loss}). 
Although such a loss supervises the target robot flow, it does not explicitly constrain these methods' intermediate representations, which may leave task-relevant scene-change information entangled with task-irrelevant visual details.
Such entanglement can reduce robustness to nuisance variations such as background appearance or lighting, which often vary across practical environments.
To mitigate this issue, we use auxiliary supervision constructed from textual descriptions of task-relevant scene changes from $\bm{I}_0$ to a goal image~$\bm{G}$.

Specifically, we construct the auxiliary targets offline as follows:
First, we provide an MLLM with $\bm{I}_0$, $\bm{G}$, and a prompt instructing it to describe task-relevant changes (e.g., robot pose changes and object displacements) while ignoring task-irrelevant visual details (see Supplementary Materials).
The MLLM then generates $N_c$ captions that describe the same manipulation from different perspectives.
Next, we embed the generated captions using Qwen3-Embedding-0.6B to obtain the target narrative representations $\{\bm{z}_{\text{\ND}}^{(i)}\}_{i=1}^{N_c}$.
Finally, we  compute the \DD loss in the narrative-representation space using $\bm{z}^{(2)}_\text{\PS}$: 
\begin{equation}
\mathcal{L}_{\text{\ND}}=
\frac{1}{N_c}
\sum_{i=1}^{N_c}
\left[
1 - \text{cossim}\left(\bm{z}^{(2)}_\text{\PS}, \bm{z}_\text{\ND}^{(i)}\right)
\right],
\end{equation}
where $\text{cossim}(\cdot, \cdot)$ denotes cosine similarity.
By aligning $\bm{z}^{(2)}_\text{\PS}$ with the multiple narrative representations, the \DD loss encourages $\bm{z}^{(2)}_\text{\PS}$ to capture scene changes that are consistently described across the generated captions.
We use the Narrative Delta loss only during training and $G$ is not required at inference time.

\subsection{Flow-as-Flow Module}
\label{4-8c}
We use Flow as Flow~\cite{faf-seno26} for generating a robot flow conditioned on $\bm{I}_0$ and $\bm{z}^{(1)}_\text{\PS}$.
The original Flow as Flow method is designed for goal-conditioned manipulation and thus cannot address language-conditioned settings. 
This limits its applicability in practical settings, where $\bm{G}$ is often unavailable before manipulation, while $\bm{\ell}$ is readily available. 
We therefore extend the method to language-conditioned settings using $\bm{z}_\text{\PS}^{(1)}$ as a language-conditioning signal.

In this module, we initialize $N$ (e.g., $10 \times 10$) points uniformly on $\bm{I}_0$, and obtain their future positions by integrating velocity fields predicted by this module $\bm{v}_\theta$ at each timestep $t$.
Let $\bm{\Xi}_{t}$ denote the ground-truth point coordinates at $t$, obtained by arranging the 2D coordinates of the $N$ points into an $N\times2$ matrix.
The corresponding velocities are denoted as $\dot{\bm{\Xi}}_t$.
We construct a target velocity field with a stabilization term as follows:
\begin{equation}
\bm{v}(\bm{\Xi}_t, \bm{X}, t) = \dot{\bm{\Xi}}_t - k(\bm{X} - \bm{\Xi}_t),
\label{eq:velocity}
\end{equation}
where $\bm{X} \sim \mathcal{N}\!\left(\bm{\Xi}_t,\ \sigma_0^2 e^{-2kt}\bm{I}\right)$ with a small constant $\sigma_0 > 0$ and a stabilization coefficient $k$.

We implement $\bm{v}_\theta$ based on the Diffusion Transformer (DiT) architecture~\cite{dit-peebles23iccv}.
Each DiT block is modulated via adaLN-Zero~\cite{dit-peebles23iccv} using a conditioning vector constructed from  $\bm{z}^{(1)}_\text{\PS}$, $\bm{I}_0$, and $t$.
Specifically, $\bm{I}_0$ and $t$ are encoded using ResNet-18~\cite{resnet-he16cvpr} and sinusoidal positional encoding, respectively, and their resulting features are added to $\bm{z}^{(1)}_\text{\PS}$ to form the conditioning vector.
At each timestep $t$, we form an input sequence $\mathcal{X}_{0:t} \in \mathbb{R}^{(t+1) \times N \times 2}$ by appending the current point coordinates $\mathcal{X}_{t}$ to the past point coordinate sequence $\mathcal{X}_{0:t-1}$.
Then, the coordinates of the $N$ points are flattened into a vector in $\mathbb{R}^{2N}$, which is used as a token.
The resulting sequence is processed by the DiT blocks, and the last token represents the velocity field.
For inference, the coordinates at $t$, denoted as $\bm{X}_t$, are obtained as follows:
\begin{equation}
\bm{X}_t = \bm{X}_0 + \int_0^t \bm{v}_\theta \left( \bm{X}_\tau, \tau \mid \bm{I}_0, \bm{z}^{(1)}_\text{\PS}, \mathcal{X}_{<\tau} \right) d\tau,
\end{equation}
where $\mathcal X_{<\tau}$ denotes the history of the point coordinates at previous timesteps.

The overall loss $\mathcal{L}$ is defined as follows:
\begin{equation}
    \mathcal{L} = \mathcal{L}_\text{FaF} + \lambda\mathcal{L}_\text{\ND},
\end{equation}
where $\lambda$ is a weighting hyperparameter, and $\mathcal{L}_\text{FaF}$ is the conditional flow matching loss~\cite{flowmatching-lipman23iclr, faf-seno26}:
\begin{equation}
\label{eq:primary_loss}
    \mathcal{L}_\text{FaF} = \mathbb{E}_{\bm{\Xi}_t, t, \bm{X}} \left[ 
    \left\| \bm{v}_\theta \left( \bm{X}, t \mid \bm{I}_0, \bm{z}^{(1)}_\text{\PS}, \bm{\Xi}_{0:t-1} \right)
    - \bm{v} \left(\bm{\Xi}_t, \bm{X}, t \right) \right\|^2 \right].
\end{equation}

%% file: section3.tex
\vspace{-1mm}
\subsection{Problem Statement}
\vspace{-2mm}
In this study, we address language-conditioned flow-based manipulation.
This task consists of two subtasks: robot flow generation and object manipulation conditioned on the generated robot flow.
In the robot flow generation subtask, given an initial image and an instruction, the model generates a robot flow, which is expected to be aligned with the ground-truth robot motion. %
Subsequently, in the object manipulation subtask, at each timestep, given the current observation and the generated robot flow, the model outputs a robot end-effector pose.
\figref{fig:eye-catch} shows an example of this task.
Given the initial image and the instruction “Pick up the gold can, and place it on the table,” the model is required to generate the corresponding robot flow.
Based on the generated robot flow, the robot then moves the gold can onto the table.

In this study, we assume that the end-effector is visible in an initial image under a fixed camera setup.
We focus on the flow generation subtask.
For the object manipulation subtask, we use a flow-conditioned manipulation policy, such as those adopted in prior methods (e.g., \cite{faf-seno26,im2flow2act-xu24corl,track2act-bharadhwaj24eccv}).

%% file: tab/4-4-model.tex
\begin{figure}[tb]
  \centering
  \includegraphics[width=\textwidth,height=6.5cm,keepaspectratio]{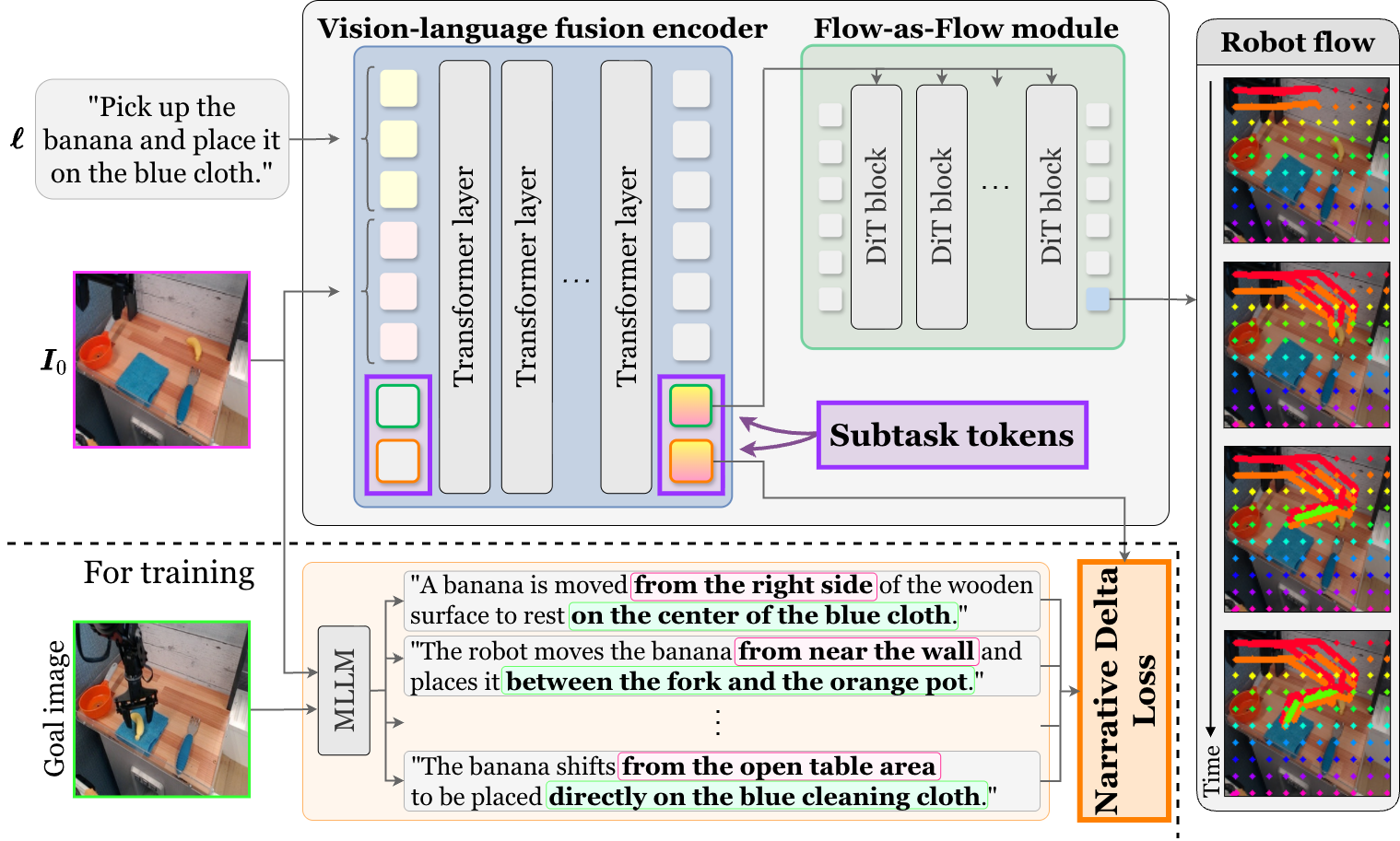}
  \caption{Architecture of \YYY. The \xxxfusion integrates the embeddings of $\bm{\ell}$ and $\bm{I}_0$  into the two \MultiXXX tokens: one for conditioning the Flow as Flow module to generate the robot flow, and the other for computing the \DD loss.
During training, the \DD loss is used as an auxiliary training objective, aligning the corresponding \XXX token with the narrative representations of task-relevant scene changes from $\bm{I}_0$ to the goal image.
}
  \label{fig:model}
\end{figure}

%% file: section5-2.tex
\section{Experiments}
\label{sec:experiments}
\subsection{Experimental Setup}
\label{experiments_setup}
\noindent\textbf{\emph{Datasets.}}
We used video clips from Fractal~\cite{rt1-brohan23rss} and Bridge V2~\cite{bridgev2-walke23corl}, both of which are standard real-robot datasets in robot learning~\cite{oxe-neill24icra, openvla-kim24corl}.
\tabref{tab:datasets} provides an overview of the datasets used in our experiments.
The two datasets were collected from different real-world robotic platforms, leading to variation in the recorded motion and visual appearance of the scenes (e.g., backgrounds and viewpoints).
For Fractal, the training, validation, and test sets contain 84,212, 1,000, and 2,000 samples, respectively. 
For Bridge V2, the corresponding sets contain 57,064, 1,000, and 2,000 samples, respectively.
We used the training sets to train the model, the validation sets to tune the hyperparameters, and the test sets to evaluate the model.

Following prior work~\cite{faf-seno26}, we curated the datasets as described below.
We set $\bm{I}_0$ as the first frame of the clip for each episode.
In addition, we set $\bm{G}$ according to the binary success label provided for each episode.
Concretely, for a successful episode, $\bm{G}$ was set as the frame at which success was determined, whereas for an unsuccessful episode, it was set as the last frame of the clip.
We used $\bm{\ell}$ corresponding to $\bm{I}_0$ when $\bm{\ell}$ was included in the original datasets; otherwise, $\bm{\ell}$ was generated by Qwen2.5-VL~\cite{qwen2.5-vl-bai25} from the given $\bm{I}_0$ and $\bm{G}$.
To obtain $\bm{\Xi}_{0:T-1}$, we used CoTracker3~\cite{cotracker3-karaev25iccv}, an off-the-shelf point-tracking method.
Here, $T$ denotes the sequence length of the robot flow.
Specifically, we initialized $N$ points uniformly on $\bm{I}_0$ and tracked their trajectories throughout the clip.

Following prior work~\cite{faf-seno26}, the datasets were preprocessed as described below.
First, all images were resized to 256 $\times$ 256 pixels.
Next, we obtained a segmentation mask of a robot end-effector from $\bm{I}_0$ for each video and incorporated it as an alpha channel of $\bm{I}_0$ in the model input.
End-effector masks were generated using Robot-SAM~\cite{roboengine-yuan25iros}.
Because the pretrained Robot-SAM often generated masks covering the entire robot arm, we fine-tuned Robot-SAM to localize only the end-effector.
Specifically, we used 300 samples from the RoboSeg~\cite{roboengine-yuan25iros} dataset with additional end-effector mask annotations.

\input{tab/5-5-datasets}

\noindent\textbf{\emph{Baselines.}}
We used representative language-conditioned robot flow generation baselines, FLIP~\cite{flip-gao25iclr} and Im2Flow2Act~\cite{im2flow2act-xu24corl}.
Both were trained on the same datasets as the proposed method.

\noindent\textbf{\emph{Metrics.}}
\label{sec:metrics}
Following prior work~\cite{faf-seno26}, we used average displacement error (ADE), final displacement error (FDE), and less than delta ratio (LTDR) as evaluation metrics, with ADE as the primary metric.
ADE was computed by averaging the Euclidean distance between $\mathcal{X}_{0:T-1}$ and $\bm{\Xi}_{0:T-1}$ over all evaluated points and all time steps, and FDE was computed by averaging the Euclidean distance between $\mathcal{X}_{T-1}$ and $\bm{\Xi}_{T-1}$ over all evaluated points at the final step.
Lower values indicate better performance for both metrics.
LTDR followed a delta-based evaluation protocol~\cite{cotracker-karaev24eccv}, computed as $\text{LTDR} = \sum_{m=1}^{M} \sum_{t=0}^{T-1} \delta_m^t / MT$. 
Here, $\delta_m^t$ denotes the fraction of points for which the Euclidean distance between $\bm{X}_t$ and $\bm{\Xi}_t$ was within $m$ pixels at each step $t$.
The LTDR score lies in $[0,1]$, with higher values indicating better performance, and we set $M=100$ in this experiment.

These metrics are standard for point tracking~\cite{cotracker-karaev24eccv} and have been adopted for robot flow evaluation~\cite{track2act-bharadhwaj24eccv, flip-gao25iclr, egoscaler-yoshida25cvpr}.
We restricted the evaluation to the end-effector region in each image rather than computing these metrics across the entire image, following prior work~\cite{faf-seno26}.
This is because background regions in robot manipulation scenes tend to remain stationary under static camera setups and would dominate the evaluation if included in the metric computation.
The end-effector region was identified by displacement-based thresholding and morphological operations, based on the assumption that points corresponding to the end-effector exhibit larger displacements than other regions.

\noindent\textbf{\emph{Implementation Details.}}
Our model had approximately 487M trainable parameters.
We trained our model on a GeForce RTX 5090 with 32GB of GPU memory.
The total training time was approximately 6 hours, and the inference time was approximately 62ms.
We used ADE as the validation metric on the validation set of the Bridge V2 dataset, computed every 2,000 steps, and the early stopping strategy was adopted if ADE did not continuously decrease for 20,000 steps. 
To generate captions for the Narrative Delta loss, we used Qwen3.5-9B~\cite{qwen35blog} and set $N_c =5$.
$\lambda$ was $0.1$ for the first 30k steps and $1.0\times10^{-5}$ thereafter.
Following prior work~\cite{track2act-bharadhwaj24eccv,faf-seno26}, we set $T=8$, $N=10\times10$, $k=1$, and $\sigma_0=0.05$.
Further details are provided in the Supplementary Materials.
\input{section6.tex}

%% file: tab/5-5-datasets.tex
\begin{table}[t]
\centering
\caption{Statistics of the datasets used in our experiments.}
\label{tab:datasets}
\makebox[\linewidth][c]{%
\scalebox{0.90}{
\setlength{\tabcolsep}{5pt}
\begin{tabular}{lcccccc}
\toprule
Dataset & Embodiment &
\makecell{Number of\\ samples} &
\makecell{Avg. frame\\rate [fps]} &
\makecell{Avg. \\duration [s]} &
\makecell{Vocabulary\\size} &
\makecell{Avg. sentence\\length [words]} \\
\midrule
Fractal~\cite{rt1-brohan23rss}
& Google Robot & 87,212 & 3.00 & 14.15 & 50 & 5.70 \\
Bridge V2~\cite{bridgev2-walke23corl}
& WidowX & 60,064 & 5.00 & 7.51 & 3,012 & 8.05 \\
\bottomrule
\end{tabular}
}%
}
\end{table}

%% file: section6.tex
\label{sec:results}
 \input{tab/6-1-quantitative}
\subsection{Quantitative Results}
\tabref{tab:quantitative} shows a quantitative comparison between the proposed method and baseline methods on the test sets of Fractal~\cite{rt1-brohan23rss} and Bridge V2~\cite{bridgev2-walke23corl}.
The proposed method achieved the best scores across the metrics on both datasets.
Specifically, it achieved ADE scores of 21.68 and 30.59 on Fractal and Bridge V2, respectively, outperforming the best baseline scores of 37.14 and 50.73 by 15.46 and 20.14 points.
Similarly, it obtained FDE scores of 31.59 and 42.42, compared with the best baseline scores of 47.74 and 68.43, respectively.
The performance differences between the proposed and baseline methods were statistically significant ($p<0.01$).

\subsection{Qualitative Results}
\input{tab/6-8-qualitative}
\figref{fig:qualitative} shows qualitative results of the proposed method compared with a baseline method, Im2Flow2Act~\cite{im2flow2act-xu24corl}. 
\figref{fig:qualitative}(i) shows a sample from the Fractal dataset, while \figref{fig:qualitative}(ii) and (iii) show samples from the Bridge V2 dataset. 
In each panel, the first row presents $\bm{\ell}$, $\bm{I}_0$, and the ground-truth robot flow.
The second and third rows show the predicted robot flows of Im2Flow2Act and the proposed method, respectively.
They are visualized at intermediate timesteps and the final timestep in temporal order.

\figref{fig:qualitative}(i) shows that the robot picked a Pepsi can from the top shelf of the fridge, where multiple cans of different colors were present in $\bm{I}_0$.
Im2Flow2Act failed to yield a robot flow toward the correct can from the phrase “Pepsi can.”
In contrast, the proposed method generated the robot flow that followed $\bm{\ell}$.

\figref{fig:qualitative}(ii) depicts a manipulation that involved picking the silver lid and then putting it on the silver pot.
Im2Flow2Act generated the robot flow that immediately moved toward the pot, without reaching the lid.
By contrast, the proposed method distinguished the lid from the pot and successfully generated the robot flow for the two-step manipulation.

In \figref{fig:qualitative}(iii), the robot moved the green cloth from the right side of the table to the left side.
Although Im2Flow2Act generated the robot flow that first grasped the correct object, the subsequent motion was in the wrong direction.
In contrast, the proposed method generated the robot flow that successfully captured the correct object and moved in the appropriate direction.

\input{tab/6-9-failure}
\figref{fig:failure} shows a failure case on Bridge V2.
The top row shows $\bm{\ell}$ and $\bm{I}_0$, and the following three rows show the robot flows of the ground truth, a baseline (Im2Flow2Act~\cite{im2flow2act-xu24corl}), and the proposed method, respectively.
In this case, $\bm{\ell}$ was “move the knob-like object to the left side of the table.”
The baseline failed to generate the robot flow toward the target object.
The proposed method incorrectly generated the flow that targeted the drawer and then opened it.
This failure was likely due to the expression “knob-like” in $\bm{\ell}$, which misled the proposed method into identifying the drawer knob as the target object.

\subsection{Ablation Studies}
\tabref{tab:ablation} shows the results of the ablation studies.
In these studies, the effects on performance of subtask tokens and the Narrative Delta loss were investigated.

\noindent\textbf{\emph{Subtask Token Ablation.}}
To investigate the benefit of using \XXX tokens, we evaluated two models against the proposed method.
Model (i) did not use any \XXX tokens; instead, it used the output tokens of the Transformer encoder to condition the Flow-as-Flow module and to compute the \DD loss.
Instead of two \XXX tokens, Model (ii) used a single shared \XXX token for both objectives.
Model (i) obtained ADE scores that were 1.52 and 2.82 points worse than those of the proposed method on Fractal and Bridge V2, respectively.
Similarly, Model (ii) yielded ADE scores 3.04 and 2.61 points worse than those of the proposed method on the two datasets, respectively.
These results indicate that the \XXX tokens allow each token to specialize for its corresponding objective, improving overall robot flow generation.

\noindent\textbf{\emph{\DD Loss Ablation.}}
To assess the contribution of the \DD loss, we removed this loss from Model (iii) and trained the model only with $\mathcal{L}_\text{FaF}$.
Model (iii) obtained ADE scores that were worse than those of the proposed method by 4.15 and 3.20 points on Fractal and Bridge V2, respectively.
This suggests that the \DD loss enhances robot flow generation by providing an auxiliary supervisory signal for learning task-relevant scene changes.
\input{tab/6-10-ablation}

%% file: tab/6-1-quantitative.tex
\begin{table}[t]
\centering
\caption{Quantitative comparison between the proposed method and baseline methods. The best scores for each metric are shown in \textbf{bold}.}
\setlength{\tabcolsep}{5pt}
\begin{tabular}{lcccccc}
\toprule & \multicolumn{3}{c}{Fractal} & \multicolumn{3}{c}{Bridge V2} \\
\cmidrule(lr){2-4} \cmidrule(lr){5-7}
Method & ADE $\downarrow$ & FDE $\downarrow$ & LTDR $\uparrow$[\%] & ADE $\downarrow$ & FDE $\downarrow$ & LTDR $\uparrow$[\%] \\
\midrule
FLIP\cite{flip-gao25iclr} & 66.17 & 87.52 & 35.69 & 50.73 & 68.43 & 47.72 \\
Im2Flow2Act\cite{im2flow2act-xu24corl}  & 37.14 & 47.74 & 60.61 & 51.48 & 70.93 & 47.97 \\
\midrule
\textbf{Ours}  & \textbf{21.68} & \textbf{31.59} & \textbf{76.29} & \textbf{30.59} & \textbf{42.42} & \textbf{66.30} \\
\bottomrule
\end{tabular}
\label{tab:quantitative}
\end{table}

%% file: tab/6-8-qualitative.tex
\begin{figure}[tb]
  \centering
  \includegraphics[width=\textwidth,height=5cm,keepaspectratio]{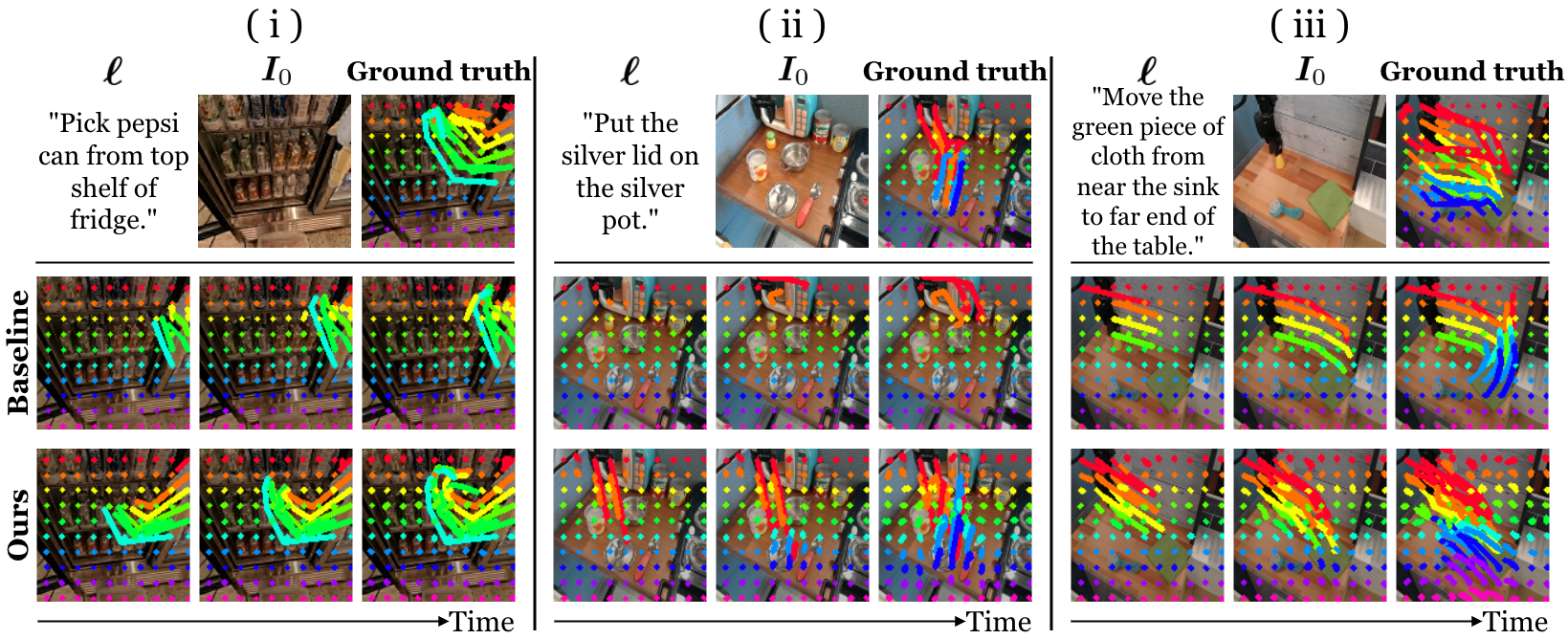}
  \caption{Qualitative comparison between \YYY and a baseline method (Im2Flow2Act~\cite{im2flow2act-xu24corl}). 
In each panel, the first row presents $\bm{\ell}$, $\bm{I}_0$, and the ground-truth robot flow.
The second and third rows show the predicted robot flows of the baseline method and the proposed method, respectively.
}
  \label{fig:qualitative}
\end{figure}

%% file: tab/6-9-failure.tex
\begin{figure}[tb]
  \centering
  \includegraphics[height=5.5cm]{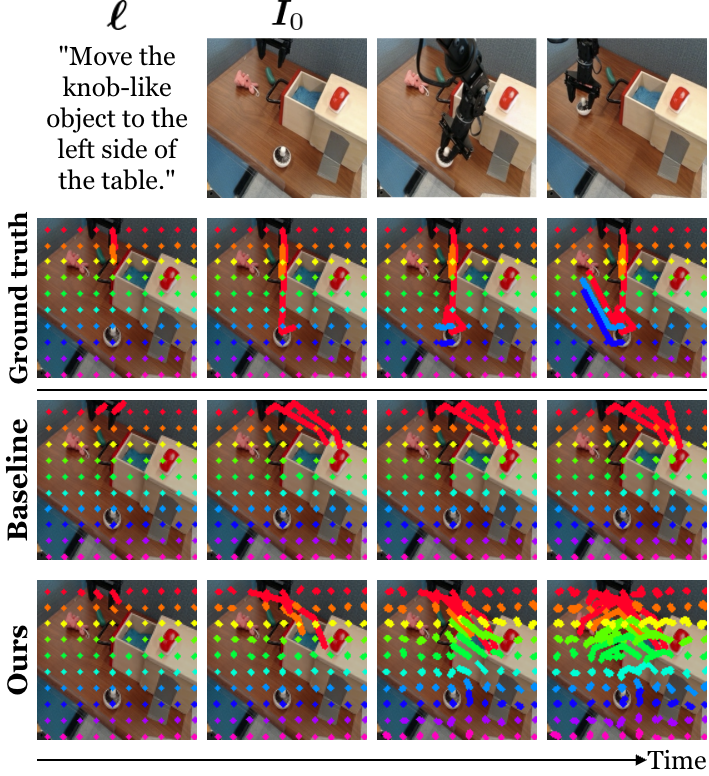}
  \caption{A failure case of \YYY on Bridge V2~\cite{bridgev2-walke23corl}.
The first row shows $\bm{\ell}$, $\bm{I}_0$, and subsequent frames sampled from the ground-truth video in temporal order. 
The other rows show the robot flows of the ground truth, a baseline (Im2Flow2Act~\cite{im2flow2act-xu24corl}), and the proposed method in temporal order, respectively.
}
  \label{fig:failure}
\end{figure}

%% file: tab/6-10-ablation.tex
\begin{table}[t]
\centering
\caption{Quantitative results of ablation studies on the \XXX tokens and the \DD loss. 
$\checkmark$ indicates the use of the \DD loss.
The best scores for each metric are shown in \textbf{bold}.
}
\resizebox{\linewidth}{!}{%
\setlength{\tabcolsep}{3pt}
\begin{tabular}{lcccccccc}
\toprule
\multirow{2}{*}[-0.7ex]{Model}
& \multirow{2}{*}[-0.7ex]{\shortstack[c]{Subtask\\[-0.2ex]tokens}}
& \multirow{2}{*}[-0.7ex]{\shortstack[c]{\DD\\[-0.2ex]loss}}
& \multicolumn{3}{c}{Fractal}
& \multicolumn{3}{c}{Bridge V2} \\
\cmidrule(lr){4-6} \cmidrule(lr){7-9}
& & &
ADE $\downarrow$ & FDE $\downarrow$ & LTDR $\uparrow$[\%]
& ADE $\downarrow$ & FDE $\downarrow$ & LTDR $\uparrow$[\%] \\
\midrule
(i)   & 0 & \checkmark  & 23.20 & 32.91 & 74.84 & 33.41 & 47.40 & 63.42 \\
(ii)  & 1 & \checkmark  & 24.72 & 33.96 & 72.89 & 33.20 & 45.30 & 63.50 \\
(iii) & 2 &             & 25.83 & 35.72 & 71.87 & 33.79 & 47.81 & 63.00 \\
\midrule
(iv) \textbf{Ours} & 2 & \checkmark & \textbf{21.68} & \textbf{31.59} & \textbf{76.29} & \textbf{30.59} & \textbf{42.42} & \textbf{66.30} \\
\bottomrule
\end{tabular}
}
\label{tab:ablation}
\end{table}

%% file: section8.tex
\section{Real-World Experiments}
\subsection{Experimental Setup}
To validate \YYY in downstream manipulation tasks, we conducted real-world experiments.

\noindent\textbf{\emph{Robot Platform.}}
\input{tab/A-1}
All of the experiments were conducted using the Human Support Robot (HSR), an 11-DoF mobile manipulator developed by Toyota Motor Corporation, as shown in \figref{fig:hsr}.
The visual inputs provided to the policy were captured with an external webcam.
Specifically, these inputs consisted of $\bm{I}_0$ and the current observation at each timestep during execution.

\noindent\textbf{\emph{Tasks.}}
We evaluated \YYY on three mobile manipulation tasks. 
\begin{enumerate}
    \item \textbf{Mobile drawer closing}: In this task, the robot is expected to close an open tabletop drawer by pushing it forward.
    The generated robot flow must provide the correct pushing direction and  an appropriate pushing stroke; otherwise, the drawer cannot be closed completely.
    \item \textbf{Mobile bin picking}:
    This task involves picking a specified object from a table, and placing it into a bin.
    Given $\bm{\ell}$, the model must identify the target object among multiple objects on the table.
    \item \textbf{Mobile cup stacking}: 
    This task requires the robot to stack a cup onto two nested cups.
    The generated robot flow must support precise alignment of the small cups; even a misalignment of a few centimeters can result in failure.
\end{enumerate}
Mobile manipulation is more challenging for a flow-conditioned policy than tabletop manipulation because the policy is required to map the generated robot flows to both end-effector and mobile-base actions.

\noindent\textbf{\emph{Baselines and Policy.}}
We used two baseline methods, Im2Flow2Act~\cite{im2flow2act-xu24corl} and FLIP~\cite{flip-gao25iclr}, which are the same baselines evaluated in Sec.~\ref{sec:experiments}.
For manipulation, we used a flow-conditioned policy following the DiT-based architecture and training setup in~\cite{faf-seno26}.
We collected an average of 30 teleoperated demonstrations for each task, and fine-tuned each model on the collected data.

\subsection{Quantitative Results}
\input{tab/A-5}
\tabref{tab:real-quantitative} shows the quantitative results of the real-world experiments.
For each method, we report success rates across 20 trials per task, along with the average rate over the three tasks.
The table shows \YYY achieved higher success rates than both baseline methods across the three tasks.
Overall, \YYY achieved an average success rate of 55\%, outperforming the strongest baseline, Im2Flow2Act~\cite{im2flow2act-xu24corl}, by 13 points.
We also report the success rates of the oracle method, in which the policy was conditioned on the corresponding ground-truth robot flows.
The average success-rate gap relative to the oracle method was 8 points for \YYY, whereas it was 21 points for Im2Flow2Act.
This smaller gap suggests that the robot flows generated by \YYY are more effective for the downstream manipulation policy than those of the baseline methods.

\subsection{Qualitative Results}
\input{tab/A-6}
\figref{fig:real-qualitative} shows the qualitative results of real-world experiments.
In each panel, the top row shows $\bm{\ell}$, $\bm{I}_0$, and the generated robot flow, while the bottom row shows the corresponding execution in temporal order.

\figref{fig:real-qualitative}(i) depicts a successful execution of the \textbf{mobile drawer closing} task.
In this case, the generated robot flow provided the correct pushing direction and an appropriate pushing stroke, enabling the robot to completely close the drawer.

\figref{fig:real-qualitative}(ii) presents a \textbf{mobile bin picking} case, where the robot should grasp the target object and place it into the designated container.
The proposed method correctly identified the target object specified by $\bm{\ell}$, and the robot completed the task using the generated robot flow.

In \figref{fig:real-qualitative}(iii), the robot successfully executed the \textbf{mobile cup stacking} task.
The generated robot flow was aligned with the positions of the target cup and the nested cups, guiding the robot to correctly perform the manipulation.

%% file: tab/A-1.tex
\begin{figure}[tb]
  \centering
  \includegraphics[width=\textwidth,height=2.5cm,keepaspectratio]{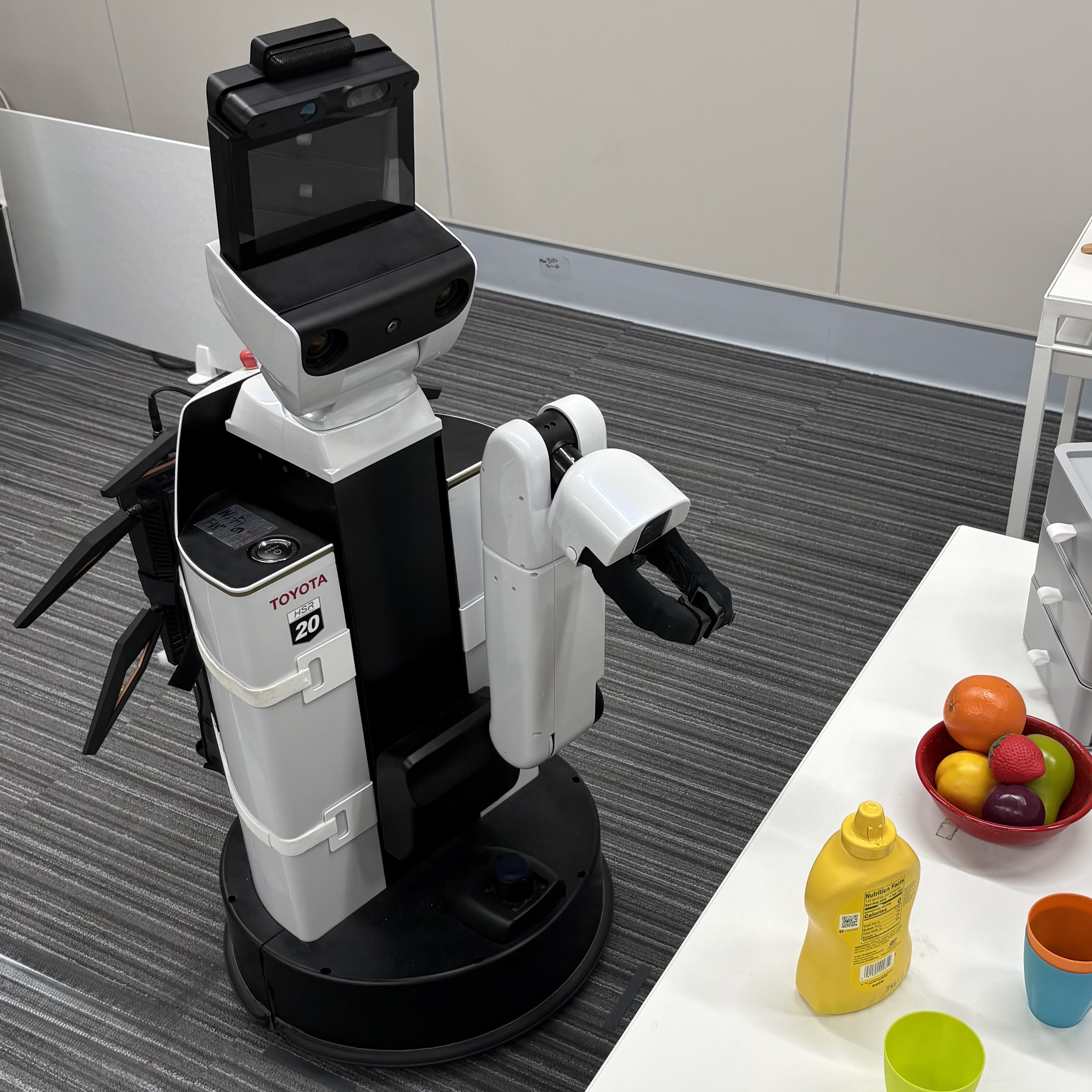}
  \caption{The Human Support Robot (HSR) used in our real-world experiments.}
  \label{fig:hsr}
\end{figure}

%% file: tab/A-5.tex
\begin{table}[t]
\centering
\caption{Quantitative results of real-world experiments. 
For each method, we report success rates across 20 trials per task, and the average  over the three tasks.
For the oracle method, the policy was conditioned on the corresponding ground-truth robot flows.
The best scores among the proposed and baseline methods are shown in \textbf{bold}.
}
\resizebox{\linewidth}{!}{%
\setlength{\tabcolsep}{5pt}
\begin{tabular}{lcccc}
\toprule
Method [\%] & Mobile drawer closing & Mobile bin picking & Mobile cup stacking & Avg. \\
\midrule
FLIP\cite{flip-gao25iclr}                 & 50 & 25 & 10 & 28 \\
Im2Flow2Act\cite{im2flow2act-xu24corl}   & 65 & 45 & 15 & 42\\
\midrule
\textbf{Ours}                   & \textbf{85} & \textbf{60} & \textbf{20} & \textbf{55}\\
\midrule
Oracle                          & 90 & 80 & 20 & 63\\
\bottomrule
\end{tabular}
}%
\label{tab:real-quantitative}
\end{table}

%% file: tab/A-6.tex
\begin{figure}[tb]
  \centering
  \includegraphics[width=\linewidth,height=3cm,keepaspectratio]{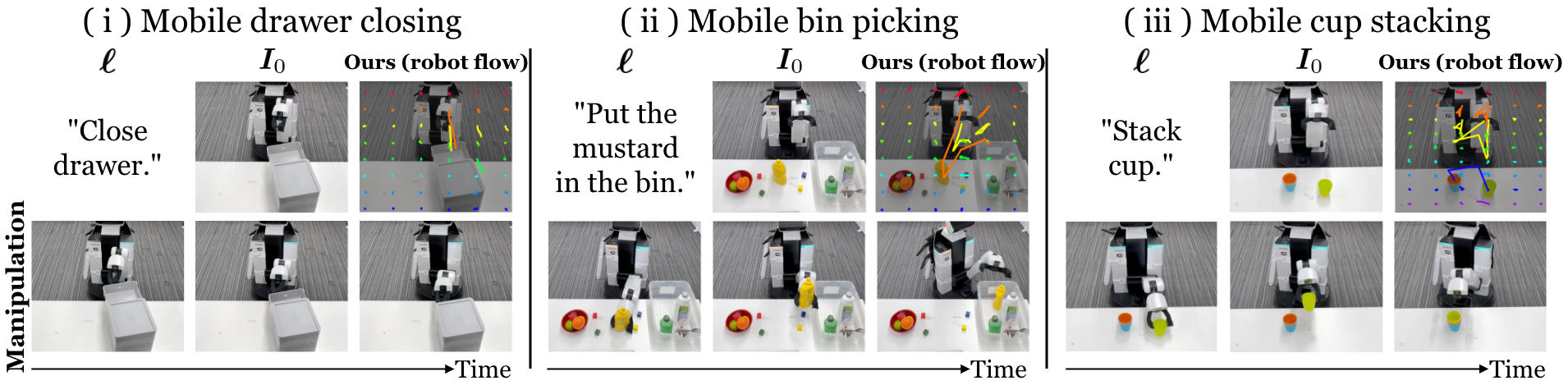}
  \caption{Qualitative results of real-world experiments. The panels show successful executions of three manipulation tasks: (i) \textbf{mobile drawer closing}, (ii) \textbf{mobile bin picking}, and (iii) \textbf{mobile cup stacking}.
In each panel, the top row shows $\bm{\ell}$, $\bm{I}_0$, and the generated robot flow, while the bottom row shows the downstream manipulation  conditioned on the robot flow.
}
  \label{fig:real-qualitative}
\end{figure}

%% file: section7.tex
\vspace{-2mm}
\section{Conclusion}
\vspace{-2mm}
In this study, we focused on language-conditioned flow-based manipulation.
The contributions of this study are as follows:
\begin{enumerate}
    \item We proposed \YYY, a language-conditioned robot flow generation method that formulates robot flows as probability velocity fields within a flow-matching framework, enabling high-quality robot flow generation.
    \item We introduced the \DD loss as an auxiliary objective that aligns the model’s intermediate representations with task-relevant scene changes. We further introduced the \MultiXXX tokens to mitigate interference between this auxiliary objective and the primary flow-generation objective.
    \item \YYY outperformed the baseline methods across standard datasets.
    \item Through real-world experiments, \YYY achieved higher success rates than the baseline methods across multiple manipulation tasks.
\end{enumerate}

Future work will include lifting 2D image-plane robot flows into 3D space by incorporating depth information. 
This could enable depth-wise manipulation, such as pulling a book straight out of a bookshelf toward the camera viewpoint. 
In this case, the book appears nearly static in the image plane, resulting in a negligible 2D flow despite substantial motion in 3D.

%% file: section9.tex
\paragraph{\textbf{Acknowledgments.}}
This work was partially supported by JST CREST and JSPS Fellows Grant Number JP26KJ1969.

%% file: supp.tex
\clearpage
\appendix

\title{NarrativeFlow: \\ Flow-Based Vision-Language-Action Model\\
Using Robot Velocity Fields\\
(Supplementary Material)}

\author{Shota Kobayashi \and Koki Seno \and Daichi Yashima \and Komei Sugiura}

\institute{Keio University, Japan
\email{\{shotakoba10267,koki.seno,ydaichi1207,komei.sugiura\}@keio.jp}
}
\titlerunning{NarrativeFlow: VLA Model using Robot Velocity Fields}
\authorrunning{S.~Kobayashi et al.}
\maketitle
\let\oldthebibliography\thebibliography
\let\endoldthebibliography\endthebibliography
\setcounter{figure}{0}
\setcounter{table}{0}
\renewcommand{\thetable}{\Alph{table}}
\renewcommand{\thefigure}{\Alph{figure}}

\section{Additional Related Work}
\noindent\textbf{\emph{Standard Datasets in Robot Learning.}}
Real-world robot manipulation data\-sets have substantially expanded in both scale and embodiment diversity.
Open X-Embodiment (OXE)~\cite{oxe-neill24icra} is a large-scale cross-embodiment dataset constructed by aggregating demonstration data from heterogeneous platforms, and it has been widely used for training vision-language-action (VLA) models (e.g.,~\cite{pi0-black25rss,pi05-black25corl,openvla-kim24corl,spatialvla-qu25rss}).
Among the datasets within OXE, Fractal~\cite{rt1-brohan23rss} and Bridge V2~\cite{bridgev2-walke23corl} provide language-annotated demonstrations for diverse manipulation tasks.
Complementing these datasets, DROID~\cite{droid-khazatsky24rss} provides large-scale manipulation demonstrations across diverse in-the-wild settings, including households and offices.
Recent efforts have further expanded the range of robot embodiments, with \mbox{AgiBot World}~\cite{agibot-bu25iros}, RoboMIND~\cite{robomind-wu25rss}, and ARIO~\cite{ario-wang24} covering diverse platforms, including humanoid robots.

\section{Implementation Details}
\label{sec:details}

\tabref{tab:hyperparameters} shows the training settings of the proposed method.
In terms of the architecture, we used four Transformer layers in the vision-language fusion encoder.
As specified in Sec.~4.1, we set $\lambda$ to a high value for the early training steps, and to a low value for the remaining steps~\cite{vipra-routray25neurips, ssl360-tran21cvpr}.
This schedule assigns a larger weight to $\mathcal{L}_\text{\ND}$ in the initial phase, grounding the model's intermediate representations in task-relevant scene changes.
In the subsequent phase, the training objective focuses on robot flow generation based on the learned scene-change prior.

In addition, for the caption generation used in the \DD loss \mbox{(Sec.~3.4),} we used the prompt shown below with a maximum token count $M$. We set $M=32$ based on preliminary caption-generation trials without a length constraint, where the generated captions were observed to fit within 32 tokens.
\input{tab/prompt}
\input{tab/5-8-hyperparameters}

\section{Preliminary Analysis of Narrative Delta Loss}
\tabref{tab:ablation-supp} shows the results of our preliminary analysis of the Narrative Delta loss.
We evaluated three design choices: the number of captions, the loss function, and the source for caption generation.

\subsection{Number of Captions}
To investigate the effect of the number of captions used in the Narrative Delta loss, we compared models with $N_c$ set to 1, 3, and 5, corresponding to Models (i), (ii), and (v), respectively.
Following the main experimental settings, each training sample was associated with five captions generated using the prompt in Sec. \ref{sec:details}.
In this analysis, each model used the first $N_c$ captions from the five captions.

As shown in \tabref{tab:ablation-supp}, the proposed model with $N_c=5$ achieved the best scores across the metrics on both datasets.
In particular, on Bridge V2, increasing $N_c$ yielded better scores across all three metrics.
These results suggest that using multiple captions is effective for the Narrative Delta loss, since it helps the model capture task-relevant semantics that are shared across different captions of the same scene change.

\subsection{Loss Function}
To assess the effect of the loss function used for the \DD loss, we compared
Model~(iii) with the proposed model. Model~(iii) replaces the cosine similarity loss in Eq.~(1) with the following L2 loss:
\begin{equation}
\label{eq:l2}
\mathcal{L}'_{\text{\ND}}
=
\frac{1}{N_c}
\sum_{i=1}^{N_c}
\bigl\|
\bm{z}^{(2)}_\text{\PS}
-
\bm{z}^{(i)}_\text{\ND}
\bigr\|_2^2 .
\end{equation}
\input{tab/ablation-supp}
As shown in Table~\ref{tab:ablation-supp}, Model~(iii) yielded worse scores than the proposed model across the three metrics on both datasets.
Since Qwen3-Embedding-0.6B used in our method \mbox{(Sec.~3.4)}  is trained with an InfoNCE-based contrastive objective with cosine similarity~\cite{qwen3embedding-zhang25}, the Narrative Delta loss with cosine similarity is expected to be more consistent with the learned embedding space.

\subsection{Source for Caption Generation}
We also evaluated Model (iv), which uses captions generated from video clips for the Narrative Delta loss, instead of captions generated from $\bm{I}_0$ and $\bm{G}$.
Specifically, for each video clip, we uniformly selected 8 frames (following $T$ in Sec.~4.1) from the segment between $\bm{I}_0$ and $\bm{G}$, including both endpoints.
We then provided the MLLM with these frames and the video-captioning prompt shown below, with $N_c=5$ and $M=32$ following Sec.~\ref{sec:details}.

\tabref{tab:ablation-supp} shows that Model (iv) performed worse than the proposed method on both datasets.
In particular, Model (iv) yielded worse ADE scores than the proposed method by 0.58 on Fractal and 1.80 on Bridge V2.
This result indicates that captions generated from $\bm{I}_0$ and $\bm{G}$ provide a more effective training signal for robot flow generation. 
We attribute this to two factors. 
First, video captions tend to emphasize aspects of the task-relevant motion that are easy to describe, rather than faithfully reflecting the full trajectory (e.g., \cite{motionbench-hong25cvpr}). 
Second, precise trajectory information is already supervised by $\mathcal{L}_\text{FaF}$ through $\bm{\Xi}_{0:T-1}$; encoding it again as a coarse text-level embedding therefore adds no effective constraint.

\input{tab/video-prompt}

\section{Additional Experiments}

\subsection{Additional Ablation Studies}
To further examine the effects of MLLM size for caption generation and continuous velocity-field modeling in the proposed method, we conducted two additional ablation studies.
\tabref{tab:additional_ablation} shows the results of these studies.

\noindent\textbf{\emph{MLLM Size for Caption Generation.}}
To assess the effect of the MLLM size for caption generation, we evaluated Model (i) using Qwen3.5-0.8B~\cite{qwen35blog}, a smaller MLLM than Qwen3.5-9B~\cite{qwen35blog} used in the proposed method. 
\tabref{tab:additional_ablation} shows that Model (i) underperformed the proposed method across the three metrics on both datasets. 
This result indicates that captions generated by a larger MLLM lead to more effective auxiliary supervision through the Narrative Delta loss, improving robot flow generation.
Nevertheless, Model (i) outperformed the baseline methods in Table~2 across the metrics.
This suggests that even with a small MLLM for caption generation, the proposed method remains sufficiently effective for robot flow generation.

\noindent\textbf{\emph{Continuous Velocity-Field Modeling.}}
To investigate the contribution of continuous velocity-field modeling in the language-conditioned setting, we evaluated Model (ii), which  generated robot flows as displacements of sparse keypoints by setting $\sigma_0 = 0$.
In this case, since sampling $\bm{X}$ deterministically yields $\bm{\Xi}_i$, robot flows are not modeled as dense velocity fields.
\tabref{tab:additional_ablation} shows that the proposed method outperformed Model (ii) across the three metrics on both datasets.
This result indicates that continuous velocity-field modeling is effective for language-conditioned robot flow generation.

\subsection{Additional Real-world Experiments}
To further evaluate the proposed method in real-world mobile manipulation beyond the three tasks in Table 4, we conducted experiments on 10 additional tasks.
\tabref{tab:additional_real-world} shows the quantitative results for a total of 13 tasks, with 260 trials per method.
The proposed method achieved a higher average success rate than the baseline methods.
These results indicate that robot flows generated by the proposed method are effective across a broader range of real-world mobile manipulation tasks.
\input{tab/additional_ablation}
\input{tab/additional_real-world_experiments}

\section{Error Analysis}
\input{tab/6-13-erroranalysis}
We define failure cases as samples in which the proposed method yielded worse ADE scores than Im2Flow2Act. 
There were 185 and 290 failure cases on Fractal and Bridge V2, respectively.
We conducted an error analysis of 100 of these cases sampled from Bridge V2, and the results are listed in \tabref{tab:erroranalysis}.
These cases can be broadly grouped into the following six categories:
\begin{itemize}
    \item \textbf{Multimodal language comprehension errors}:
This refers to cases where the model incorrectly interpreted $\bm{\ell}$ and $\bm{I}_0$; for example, misunderstanding the target object.
A representative example was a case in which the model generated $\mathcal{X}_{0:T-1}$ that corresponded to grasping an eggplant, although $\bm{\ell}$ was ``place the spatula just behind the eggplant.''
    \item \textbf{Intermediate trajectory deviations}:
This category includes cases in which the intermediate trajectories of $\mathcal{X}_{0:T-1}$ deviated substantially from those of $\bm{\Xi}_{0:T-1}$, while $\mathcal{X}_{0:T-1}$ corresponded to manipulating the correct object toward the appropriate target location.
    \item \textbf{Terminal position errors}: 
This category refers to cases in which the terminal position of the target object deviated from the ground-truth terminal position, while the intermediate trajectories of $\mathcal{X}_{0:T-1}$ remained close to those of  $\bm{\Xi}_{0:T-1}$.
As in the \textbf{intermediate trajectory deviations} category, the target object was correctly identified and manipulated.
    \item \textbf{Ambiguous language instructions}: 
This category includes cases in which $\bm{\ell}$ was ambiguous.
For example, when two block towers were visible in $\bm{I}_0$ and $\bm{\ell}$ was ``move the block from one tower to the other,'' $\bm{\ell}$ did not uniquely identify which block should be manipulated.
This ambiguity left multiple candidate objects equally plausible; consequently, some degree of randomness in object selection was unavoidable.
    \item \textbf{Target object occlusion}: 
This category refers to cases in which the target object was heavily occluded in $\bm{I}_0$.
This lack of visibility prevented the model from identifying it.
    \item \textbf{Annotation errors}: 
This category covers cases in which either $\bm{\Xi}_{0:T-1}$ or $\bm{\ell}$ was unreliable. 
These cases involved tracking failures in CoTracker3~\cite{cotracker3-karaev25iccv} or mismatches between $\bm{\ell}$ and the manipulation in the source demonstration.
Among the 17 cases in this category, 13 involved errors in the human annotation of $\bm{\ell}$ in the Bridge V2 dataset.
\end{itemize}

\tabref{tab:erroranalysis} shows that the main challenge was multimodal language comprehension.
A possible solution is to introduce an object-centric region proposal module, which first extracts candidate object regions from $\bm{I}_0$ and then selects the region most relevant to $\bm{\ell}$ before robot flow generation.
By explicitly identifying the target object, this module is expected to mitigate failures caused by incorrect target-object identification.

%% file: tab/prompt.tex
\TColorBox{Full prompt for the Narrative Delta loss}{
You are given two images from the same robot manipulation episode.
The first image is the initial frame, and the second image is the final frame.\\

Describe only the meaningful visual difference between the initial and final frames.
Focus on object motion, hand or robot motion, state changes, spatial relations, and changes caused by the action.
Do not mention unchanged objects or background unless needed to clarify the change.\\

Write \texttt{\{$N_c$\}} non-duplicate descriptions for these same two frames.
Each description should use different wording and emphasize a different meaningful aspect of the visual change when possible.
Each description must be at most \texttt{\{$M$\}} tokens.
Each description must be one short sentence.\\

Return only a valid JSON object.
The keys must be description ids from 0 to \texttt{\{$N_c - 1$\}}, and each value must be one concise English sentence.
Do not explain your reasoning.
Do not write drafts or analysis.
Do not include markdown fences or any extra text.
Only produce the final JSON object.
}

%% file: tab/5-8-hyperparameters.tex
\begin{table}[t]
\centering
\caption{Training settings of the proposed method.}
\begin{tabular}{lc}
\toprule
Optimizer & AdamW \\
Batch size & 128 \\
Learning rate & $1.0 \times 10^{-4}$ \\
LR schedule & Cosine decay \\
Warmup steps & 9,000 \\
Training steps & 300,000 \\
Weight decay & 0.01 \\
EMA decay & 0.9999 \\
\bottomrule
\end{tabular}
\label{tab:hyperparameters}
\end{table}

%% file: tab/ablation-supp.tex
\begin{table}[t]
\centering
\caption{Quantitative results of preliminary analysis of the Narrative Delta loss.
$\bm{I}_0$/$\bm{G}$ indicates that captions were generated from $\bm{I}_0$ and $\bm{G}$, while Video indicates that captions were generated from multiple frames sampled from the video clip.
The best scores for each metric are shown in \textbf{bold}.
}
\resizebox{\linewidth}{!}{%
\setlength{\tabcolsep}{3pt}
\begin{tabular}{lccccccccc}
\toprule
\multirow{2}{*}[-0.7ex]{Model}
& \multirow{2}{*}[-0.7ex]{\shortstack[c]{$N_c$}}
& \multirow{2}{*}[-0.7ex]{\shortstack[c]{Loss\\[-0.2ex]function}}
& \multirow{2}{*}[-0.7ex]{\shortstack[c]{Caption\\[-0.2ex]source}}
& \multicolumn{3}{c}{Fractal}
& \multicolumn{3}{c}{Bridge V2} \\
\cmidrule(lr){5-7} \cmidrule(lr){8-10}
& & & &
ADE $\downarrow$ & FDE $\downarrow$ & LTDR $\uparrow$[\%]
& ADE $\downarrow$ & FDE $\downarrow$ & LTDR $\uparrow$[\%] \\
\midrule
(i)   & 1 & Cosine similarity & $\bm{I}_0/\bm{G}$ & 21.70 & 31.68 & 76.26 & 31.24 & 43.64 & 65.42 \\
(ii)  & 3 & Cosine similarity & $\bm{I}_0/\bm{G}$ & 21.71 & 31.68 & 76.24 & 30.63 & 42.80 & 66.09 \\
(iii) & 5 & L2     & $\bm{I}_0/\bm{G}$ & 27.00 & 36.26 & 70.36 & 34.98 & 48.28 & 61.69 \\
(iv)  & 5 & Cosine similarity & Video & 22.26 & 31.76 & 75.63 & 32.39 & 45.60 & 64.70 \\
\midrule
(v) \textbf{Ours} & 5 & Cosine similarity & $\bm{I}_0/\bm{G}$ & \textbf{21.68} & \textbf{31.59} & \textbf{76.29} & \textbf{30.59} & \textbf{42.42} & \textbf{66.30} \\
\bottomrule
\end{tabular}
}
\label{tab:ablation-supp}
\end{table}

%% file: tab/video-prompt.tex
\TColorBox{Full prompt to generate captions from videos}{
You are given frames sampled chronologically from a robot manipulation video.
Describe the primary action visible across the frames. 
Where relevant, include the manipulated object's initial position, trajectory, and final position, as well as any meaningful state changes and notable changes in object shape, configuration, or spatial relationships. 
Use clear, natural spatial language grounded in what is visible, drawing on directional and positional terms. 
Keep descriptions concise but specific, and omit background elements that remain unchanged throughout.

Do not explain your reasoning or produce any intermediate analysis. 

Write \texttt{\{$N_c$\}} non-duplicate descriptions of the video.
Vary the wording across descriptions: use different expressions for positions, regions, and state changes rather than repeating the same phrasing.
Each description must be a single short sentence within \texttt{\{$M$\}} tokens.
Return a valid JSON object with integer keys from 0 to \texttt{\{$N_c - 1$\}}, each mapping to one description string. Do not include markdown fences or any surrounding text.
Output only the final JSON object.
}

%% file: tab/additional_ablation.tex
\begin{table}[t]
\caption{Quantitative results of the additional ablation studies.}
\label{tab:additional_ablation}
\centering

\resizebox{\columnwidth}{!}{%
\setlength{\tabcolsep}{6pt}
\renewcommand{\arraystretch}{1.0}
\begin{tabular}{lcccccccc}
\toprule
\multirow{2}{*}{\makecell{Model}}
& \multirow{2}{*}{\makecell{Caption \\ generator}}
& \multirow{2}{*}{\makecell{$\sigma_0$}}
& \multicolumn{3}{c}{Fractal}
& \multicolumn{3}{c}{Bridge V2} \\
\cmidrule(lr){4-6}
\cmidrule(lr){7-9}
& & & ADE$\downarrow$ & FDE$\downarrow$ & LTDR$\uparrow$[\%]
& ADE$\downarrow$ & FDE$\downarrow$ & LTDR$\uparrow$[\%] \\
\midrule
(i) & Qwen3.5-0.8B & 0.05 & 26.40 & 36.59 & 71.07 & 32.72 & 44.60 & 63.89 \\
(ii) & Qwen3.5-9B & 0 & 26.43 & 36.83 & 70.86 & 34.35 & 48.01 & 62.28 \\
(iii) \textbf{Ours} & Qwen3.5-9B & 0.05
& \textbf{21.68} & \textbf{31.59} & \textbf{76.29}
& \textbf{30.59} & \textbf{42.42} & \textbf{66.30} \\
\bottomrule
\end{tabular}%
}

\end{table}

%% file: tab/additional_real-world_experiments.tex
\begin{table*}[t]
\captionof{table}{Additional quantitative results on real-world experiments across 13 manipulation tasks.}
\label{tab:additional_real-world}

\resizebox{\textwidth}{!}{%
\begin{tabular}{lcccccccccccccc}
\toprule
Method [\%]
& \makecell{Mobile \\ bin\\pushing}
& \makecell{Mobile \\ chair\\pushing}
& \makecell{Mobile \\ drawer\\closing}
& \makecell{Mobile \\ box\\closing}
& \makecell{Mobile \\ towel\\taking}
& \makecell{Mobile \\ bin\\picking}
& \makecell{Mobile \\ object\\placing}
& \makecell{Mobile \\ table\\bussing}
& \makecell{Mobile \\ laptop\\closing}
& \makecell{Mobile \\ water\\pouring}
& \makecell{Mobile \\ drawer\\opening}
& \makecell{Mobile \\ cup\\stacking}
& \makecell{Mobile \\ block\\stacking}
& Avg. \\
\midrule

FLIP\cite{flip-gao25iclr}
& 45 & 55 & 50 & 60 & 50 & 25 & 15 & 15 & 25 & 5
& 10 & 10 & 5 & 28 \\

Im2Flow2Act\cite{im2flow2act-xu24corl}
& 70 & 70 & 65 & 65 & 60 & 45 & 50 & 40 & 45 & 20
& \textbf{20} & 15 & \textbf{20} & 45 \\

\midrule
\textbf{Ours}
& \textbf{90}
& \textbf{90}
& \textbf{85}
& \textbf{85}
& \textbf{70}
& \textbf{60}
& \textbf{65}
& \textbf{70}
& \textbf{50}
& \textbf{35}
& \textbf{20}
& \textbf{20}
& \textbf{20}
& \textbf{58} \\

\bottomrule
\end{tabular}%
}
\end{table*}

%% file: tab/6-13-erroranalysis.tex
\begin{table}[t]
\centering
\caption{Categorization of failure cases.}
\label{tab:failure_cases}
\begin{tabular}{lc}
\toprule
Error category & \# Errors \\
\midrule
Multimodal language comprehension errors & 30 \\
Intermediate trajectory deviations & 22 \\
Terminal position errors & 14 \\
Ambiguous language instructions & 9 \\
Target object occlusion & 8 \\
Annotation errors & 17 \\
\midrule
Total & 100 \\
\bottomrule
\end{tabular}
\label{tab:erroranalysis}
\end{table}